\documentclass{article}

\PassOptionsToPackage{numbers, compress}{natbib}
\usepackage[final]{nips_2016}

\usepackage[utf8]{inputenc} 
\usepackage[T1]{fontenc}    
\usepackage[breaklinks=true,colorlinks,bookmarks=false]{hyperref}       
\usepackage{url}            
\usepackage{booktabs}       
\usepackage{microtype}      

\usepackage{amsmath,amssymb}
\usepackage{graphicx, subfig}
\usepackage[bold]{hhtensor}

\title{Dense CNN Learning with Equivalent Mappings}

\author{
  Jianxin Wu \qquad Chen-Wei Xie \qquad Jian-Hao Luo\\
  National Key Laboratory for Novel Software Technology, Nanjing University \\
  163 Xianlin Avenue, Qixia District, Nanjing 210023, China \\
  \texttt{\{wujx,xiecw,luojh\}@lambda.nju.edu.cn}
}

\begin{document}

\maketitle

\begin{abstract}
Large receptive field and dense prediction are both important for achieving high accuracy in pixel labeling tasks such as semantic segmentation. These two properties, however, contradict with each other. A pooling layer (with stride 2) quadruples the receptive field size but reduces the number of predictions to 25\%. Some existing methods lead to dense predictions using computations that are not equivalent to the original model. In this paper, we propose the equivalent convolution (eConv) and equivalent pooling (ePool) layers, leading to predictions that are both dense and equivalent to the baseline CNN model. Dense prediction models learned using eConv and ePool can transfer the baseline CNN's parameters as a starting point, and can inverse transfer the learned parameters in a dense model back to the original one, which has both fast testing speed and high accuracy. The proposed eConv and ePool layers have achieved higher accuracy than baseline CNN in various tasks, including semantic segmentation, object localization, image categorization and apparent age estimation, not only in those tasks requiring dense pixel labeling.
\end{abstract}

\section{Introduction}

Large receptive filed and densely repeated nonlinear mapping (dense predictions) are both important properties of many successful deep convolutional neural network (CNN) models. These two properties are, however, not compatible with each other in classic CNN models.

For example, in the VGG-16 CNN model~\cite{vi:Simonyan2015}, the output after the \texttt{pool5} layer is a $7 \times 7 \times 512$ tensor, each element in this order 3 tensor is computed from a $212 \times 212$ receptive field in the input image (which is $224 \times 224$ in size). In CNN models for semantic segmentation, the last layer's receptive field is even larger. For example, every output of the fully convolutional network (FCN)~\cite{vi:Long2015_FCN} is computed from a $404 \times 404$ receptive field. A larger receptive field incorporates more information from the input image and often leads to more accurate recognition or segmentation results. The increase of receptive field size is mainly achieved by the pooling layers (with stride $>1$) in current CNN models. A $2 \times 2$ pooling layer with stride 2 will quadruple the receptive field size, and a $3 \times 3$ convolution with stride 1 will increase the receptive field size by 2 both horizontally and vertically.

When the input image size is larger than one layer's receptive field size, the same nonlinear mapping is applied to different regions in the input images. For example, suppose for a large input image, the \texttt{pool5} layer output $(i,j,:)$ is computed by a nonlinear mapping $f$ using a $212 \times 212$ image window whose top left corner is $(x,y)$ . Then, the \texttt{pool5} layer output $(i+1,j,:)$ is computed by the \emph{same} nonlinear mapping $f$ using the image window at $(x+32,y)$ (if padding is used, the exact receptive field might change). Applying the same mapping many times to obtain dense predictions is important to improve CNN accuracies too. For example, CNN models for semantic segmentation apply the same nonlinear mapping to different regions of an input image to obtain pixel labeling results.

Large receptive field size and dense prediction, however, contradict with each other. One pooling layer will in general quadruple the receptive filed size, but reduces the number of predictions to a quarter of the number before the pooling layer. In the FCN model, the number of nonlinear mapping outputs (predictions) is reduced to only $16 \times 16$ for a $500 \times 500$ input image ($700 \times 700$ after padding), which has to be upsampled to roughly 32 times to compare with the $500 \times 500$ groundtruth. In other words, approximately every 1000 pixels is given only one prediction by the FCN model before the upsampling. The upsampling step is an interpolation, which does not increase the real number of predictions.

Many methods have been proposed to make the nonlinear mapping $f$ be computed more times, i.e., to make \emph{denser} predictions, mainly in object detection and semantic segmentation tasks. These methods are, however, \emph{not dense enough} or the new dense nonlinear mapping is \emph{not equivalent to $f$}.

Suppose $f(\vec{x};\vec{w})$ is a nonlinear mapping specified by parameters $\vec{w}$, which is implemented by multiple layers in a CNN. For another nonlinear mapping applied to the input $\vec{x}$, we say $f$ and $g$ is \textbf{\emph{equivalent}} if $f$ and $g$ will \emph{lead to the same output if the parameters are the same}, i.e., $f(\vec{x};\vec{w})=g(\vec{x};\vec{w})$. Obviously, an equivalent nonlinear mapping $g$ has to have the same (or equivalent) network architecture as $f$, and the parameters of $g$ must have the same size and format as that of $f$. Compared to non-equivalent ones, an equivalent mapping has the following advantages:
\begin{description}
 \item[Transform existing models to have dense predictions.] Many existing CNN models can be directly utilized, e.g., as an excellent initialization for fine-tuning the new dense prediction models. A simple approach for dense equivalent mapping is to set the stride of the \emph{last} pooling layer to 1 or ``shift and merge''~\cite{vi:Sermanet2014_Overfeat,lr:Pinheiro2014_RCNN,lr:Wu2016_SemanticSeg}. However, if the stride in the last pooling layer was $s$ in the mapping $f$, setting the stride to 1 in the new mapping $g$ only increases the number of predictions by $s^2$ times. Setting the stride to 1 for more pooling layers will increase the number of predictions, but will lead to non-equivalent mappings~\cite{vi:Chen2015_DeepLab}.
 \item[Inverse transfer a slow dense net's parameters back to improve the fast original CNN.] More importantly, a dense equivalent mapping $g$ allows learning a better set of parameters $\hat{\vec{w}}$, because more training instances, estimated probabilities and information in the groundtruth are utilized in learning the parameters of $g$. Because of the equivalence, we can \emph{directly apply the parameters $\hat{\vec{w}}$ to $f$}, and $f(\vec{x};\hat{\vec{w}})$ is \emph{a better CNN model than the original CNN $f(\vec{x};\vec{w})$}! Dense computations also means that $g$ has much higher computational cost than $f$. However, $f(\vec{x};\hat{\vec{w}})$ has the same computational complexity as the original model $f(\vec{x};\vec{w})$, but its accuracy is close to that of the dense mapping $g(\vec{x};\hat{\vec{w}})$, as will be shown by our experiments. To the best of our knowledge, this paper is the first to transfer the parameters of a dense computation model back to the original model, which strikes a balance between high accuracy and affordable computational cost.
 \item[Modular architecture and wider application domains.] We achieve dense predictions by introducing two new layers into existing CNN models: the equivalent convolution (eConv) and equivalent pooling (ePool) layers, and a new hyper-parameter: equivalent stride (eST). The dilated convolution layer~\cite{vi:Chen2015_DeepLab,lr:Yu2016_DilationConv} effectively increases the receptive field size, which has achieved high accuracy in semantic segmentation. However, dilated convolution leads to non-equivalent mappings. The dilated convolution is the same as our eConv layer, but we can obtain equivalent mappings if the proposed ePool layers replace ordinary pooling layers. With the combination of eConv and ePool, we can produce dense predictions or estimations in CNN models for semantic segmentation, recognition (classification), localization and age estimation (regression) tasks, and our experiments show that these dense predictions are effective in improving the accuracy of all these models. Turning eConv and ePool layers back into ordinary convolution and pooling layers (but with their better parameters) lead to CNN models with higher accuracy but the same complexity as their respective original models.
\end{description}

The rest of this paper is organized as follows. We introduce the details of eConv and ePool layers in Section~\ref{sec:eConvePool}, which also includes a comparison of these layers with related works. Empirical validation results and relevant discussions are presented in Section~\ref{sec:exp}. Section~\ref{sec:conclusion} ends this paper with concluding remarks and discussions for future work.

\section{Dense equivalent computation using eConv and ePool layers} \label{sec:eConvePool}

\begin{figure}
 \centering
 \subfloat[Illustration of the classic convolution and pooling layers] { \includegraphics[width=0.98\textwidth]{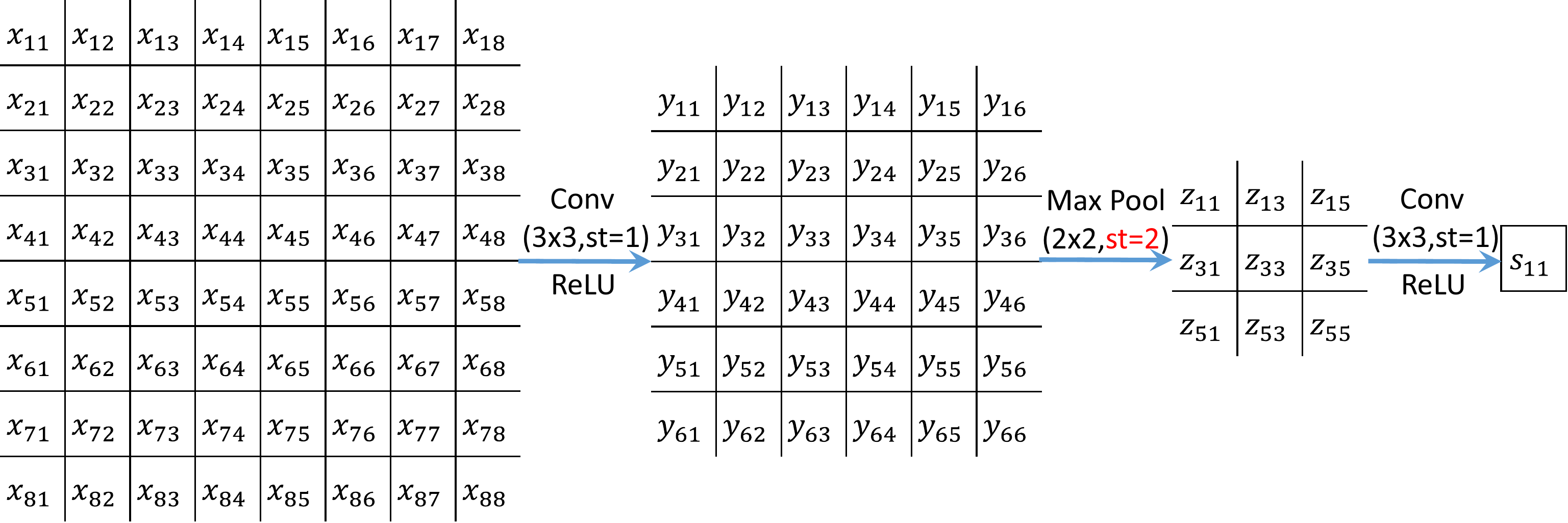} \label{fig:illu_classic} } \\
 \subfloat[After a classic pooling layer with stride 1, the equivalent convolution layer requires an equivalent stride 2 to obtain the same (equivalent) computations as the network in Figure~(\ref{fig:illu_classic})] { \includegraphics[width=0.98\textwidth]{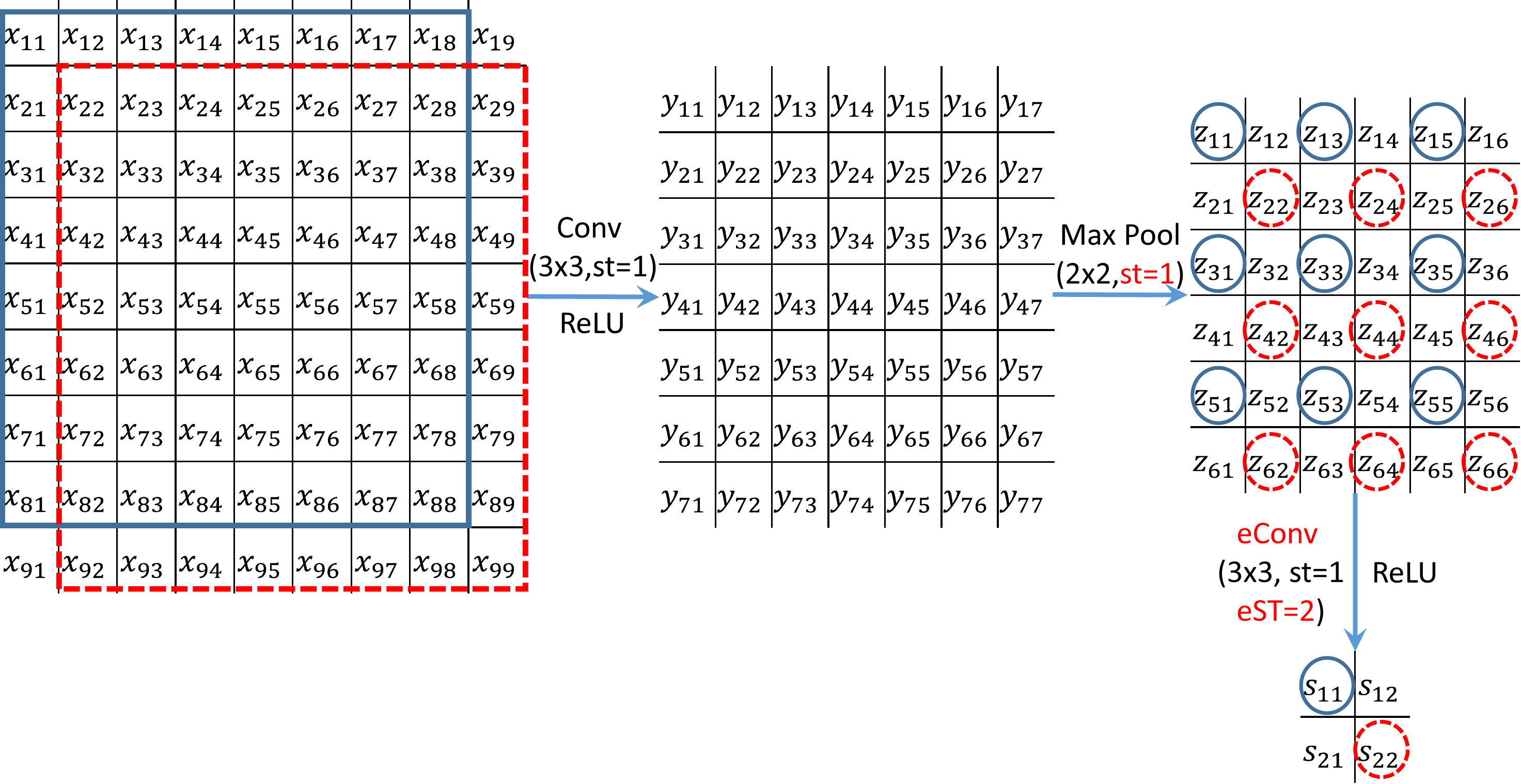} \label{fig:illu_eConv} } \\
 \subfloat[The equivalent pooling layer (whose eST=2) increase the equivalent stride for subsequent layers to eST=4] { \includegraphics[width=0.98\textwidth]{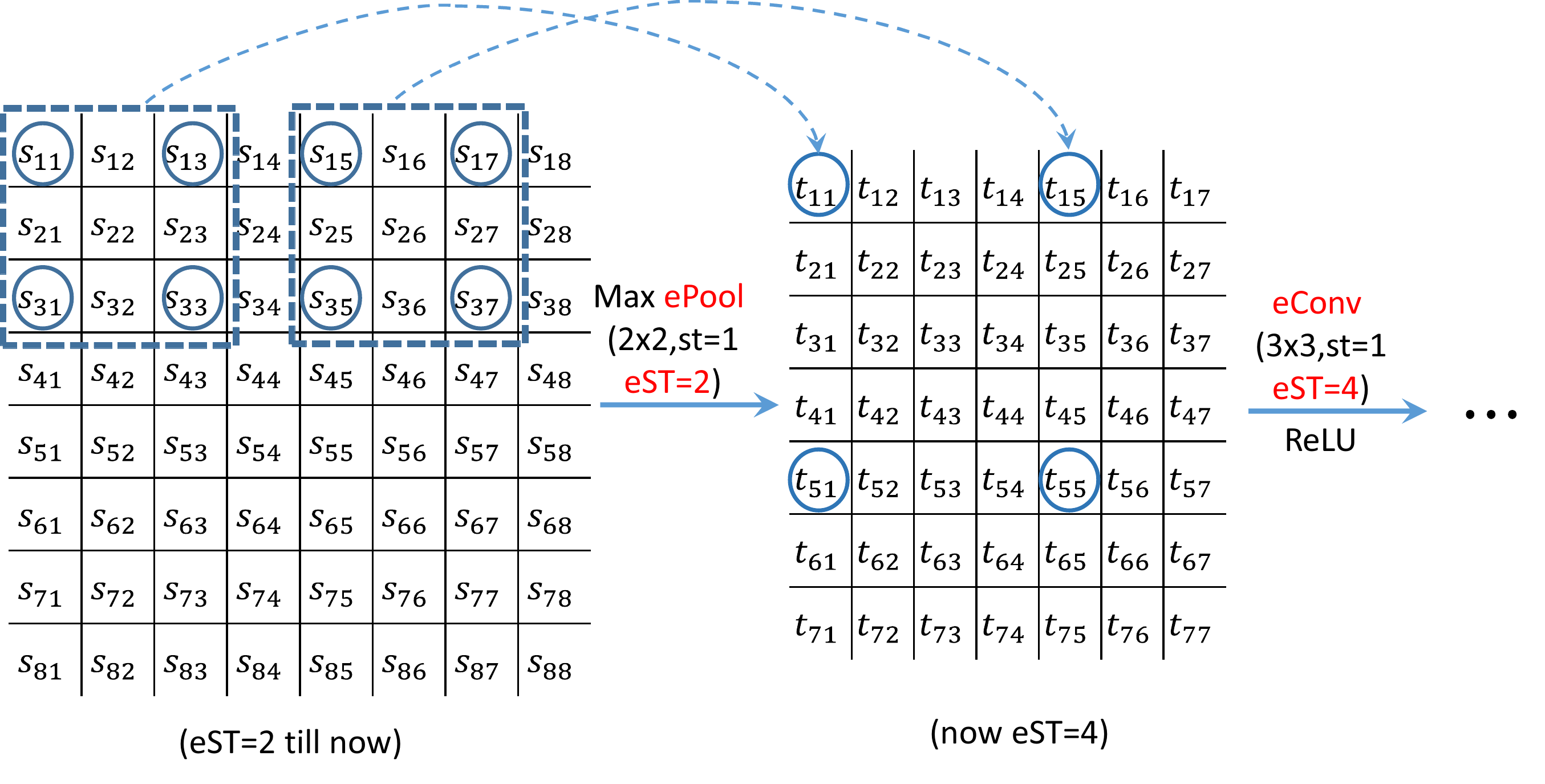} \label{fig:illu_ePool}}
 \caption{Illustration of the equivalence between classic convolution/pooling layers with eConv/ePool layers. This figure is best viewed in color.} \label{fig:illu}
\end{figure}

Equation~\ref{eqn:struct1} illustrates a classic CNN model with convolution, pooling and ReLU layers, in which the variables in different layers are denoted as $x$, $y$, $z$, $s$ and $t$, respectively. Classic convolution and pooling layers are illustrated in Figure~\ref{fig:illu_classic}. Without padding, a $3 \times 3$ convolution layer reduces the output size by 2 and a pooling layer halves the output size in both directions. In Figure~\ref{fig:illu}, we use the coordinates of the top-left element involved in a convolution or pooling operation to denote the output, e.g., $\left[\begin{smallmatrix} y_{55} & y_{56} \\ y_{65} & y_{66} \end{smallmatrix}\right]$ is max pooled to form $z_{55}$ in Figure~\ref{fig:illu_classic}.

\begin{align}
 \boxed{x} & \xrightarrow[\text{\shortstack{ReLU\\Layer 1}}]{\text{Conv (3x3, st=1)}} 
 \boxed{y} \xrightarrow[\text{\shortstack{(2x2, st=2)\\Layer 2}}]{\text{Max Pool}}
 \boxed{z} \xrightarrow[\text{\shortstack{ReLU\\Layer 3}}]{\text{Conv (3x3, st=1)}\phantom{e}}
 \boxed{s} \xrightarrow[\text{\shortstack{\phantom{T=2}(2x2, st=2)\phantom{,eS}\\Layer 4}}]{\text{Max Pool}}
 \boxed{t} \xrightarrow[\text{\shortstack{ReLU\\Layer 5}}]{\text{\phantom{e}Conv (3x3, st=1)}} \cdots \label{eqn:struct1}\\
 \boxed{x} & \xrightarrow[\text{\shortstack{ReLU\\Layer 1}}]{\text{Conv (3x3, st=1)}} 
 \boxed{y} \xrightarrow[\text{\shortstack{(2x2, \textcolor{red}{st=1})\\Layer 2$'$}}]{\text{Max Pool}}
 \boxed{z} \xrightarrow[\text{\shortstack{\textcolor{red}{eST=2}, ReLU\\Layer 3$'$}}]{\text{\textcolor{red}{eConv} (3x3, st=1)}}
 \boxed{s} \xrightarrow[\text{\shortstack{(2x2, st=\textcolor{red}{1}), \textcolor{red}{eST=2}\\Layer 4$'$}}]{\text{Max \textcolor{red}{ePool}}}
 \boxed{t} \xrightarrow[\text{\shortstack{\textcolor{red}{eST=4}, ReLU\\Layer 5$'$}}]{\text{\textcolor{red}{eConv} (3x3, st=1)}} \cdots \label{eqn:struct2}
\end{align}

Equation~\ref{eqn:struct2} shows the structure of a dense equivalent mapping for the CNN specified by Equation~\ref{eqn:struct1}. The stride of pooling layers are all set to 1 (layer $2'$ and $4'$). After the first pooling layer, convolution and pooling layers are changed to eConv and ePool layers with appropriate eST values. The equivalent stride is 1 initially, but is multiplied by $s$ after we change the stride of a pooling layer from $s$ to 1.

\subsection{The equivalent convolution and equivalent pooling layer}

Figure~\ref{fig:illu_eConv} illustrates two changes between the architectures in Equation~\ref{eqn:struct2} and~\ref{eqn:struct1}: setting the stride to 1 in one pooling layer, and an equivalent convolution layer (eConv) after it.

The output of a stride 2 pooling has only half size as its input in both directions, but a stride 1 pooling almost does not reduce the output size. Hence, denser predictions are available by reducing the stride size. However, as shown in Figures~\ref{fig:illu_classic} and~\ref{fig:illu_eConv}, further applying a convolution layer will lead to non-equivalent mappings. In Figure~\ref{fig:illu_classic}, $s_{11}$ is convolved using $z_{ij}$, $i,j\in\{1,3,5\}$, but a classic convolution operator applied to Figure~\ref{fig:illu_eConv} will generate $s_{11}$ using $z_{ij}$, $i,j\in\{1,2,3\}$!

The solution to correct this non-equivalence is to add an equivalent stride (eST) in the next convolution to make them equivalent. Let $\vec{z} \in \mathbf{R}^{H \times W \times D}$ be the input tensor to the convolution, $\vec{s}$ be the convolution result, and $\vec{f}\in\mathbb{R}^{h \times w \times D}$ be one convolution kernel, the equivalent convolution is simply
\begin{equation}
 s_{ij} = \sum_{1 \le i' \le h}\sum_{1 \le j' \le w}\sum_{1 \le d \le D} f_{i',j',d} \times z_{i+(i'-1) \times eST, j+(j'-1) \times eST,d} \,. \label{eqn:eConv}
\end{equation}
Because the current eST is 2, it is clear that now the $s_{11}$ element will have the same value (i.e., be equivalent) in Figures~\ref{fig:illu_classic} and~\ref{fig:illu_eConv}, since it is convolved using these $z$ elements in blue circles. The equivalence holds for all elements that appear in Figure~\ref{fig:illu_classic}, i.e., $s_{ij}$ when both $i$ and $j$ are odd numbers. The other $s_{ij}$ elements in Figure~\ref{fig:illu_eConv} are the new dense prediction results. For example, $s_{22}$ (shown in the red circle) is computed from the receptive field $x_{ij}$, $2 \le i,j \le 9$, which is the red square in the input image.

The subsequent layer is an equivalent pooling layer (illustrated in Figure~\ref{fig:illu_ePool}). In the original network, $s_{11}$, $s_{13}$, $s_{31}$ and $s_{33}$ are max-pooled to form $t_{11}$ (cf. Figure~\ref{fig:illu_classic}, some elements are not shown in that figure). A max pooling using an equivalent stride of 2 (cf. the circled elements in the blue dashed square in Figure~\ref{fig:illu_ePool}) produces $t_{11}$ using exactly the same computations. An equivalent max pooling with size $h \times w$ is defined by
\begin{equation}
 t_{i,j,d} = \max_{\substack{1 \le i' \le h\\1 \le j' \le w}} s_{i+(i'-1) \times eST,j+(j'-1) \times eST,d} \,.\label{eqn:ePool}
\end{equation}
The two dashed blue squares denote two max-pooling operations performed in the original network. Hence, after this equivalent pooling layer, the equivalent stride should be 4 (multiplied by 2) to maintain the equivalence.

As illustrated in Figure~\ref{fig:illu}, the eConv and ePool layers (when equipped with appropriate eST values for different layers) maintain the equivalent mapping property with the original network, but produces much denser predictions (outputs). It is easy to observe that ReLU, batch normalization and fully connected layers (when implemented as convolution/eConv) will not invalidate the equivalence between the nonlinear mappings. Hence, the proposed eConv and ePool layers can turn many existing CNN models into an equivalent model and can output dense predictions.

We implement the eConv layer by rewriting the \texttt{im2col} and \texttt{col2im} functions in Caffe~\cite{vi:Jia2014}, and write a new ePool layer. These implementations enable both the forward pass and error back propagation in CNN learning for the new dense prediction net.

\subsection{Related methods}

The proposed equivalent convolution layer has the same computations (Equation~\ref{eqn:eConv}) as the dilated convolution kernel~\cite{lr:Yu2016_DilationConv,vi:Chen2015_DeepLab}. However, without the proposed equivalent pooling layer (Equation~\ref{eqn:ePool}), dilated convolution leads to a non-equivalent computation to the original model. Dilated convolution can incorporate information from pixels in a large context, but only from a sparse subset of pixels in the receptive filed. Dilated convolution has been successfully applied in semantic segmentation~\cite{lr:Yu2016_DilationConv,vi:Chen2015_DeepLab,vi:Ondruska2016_RNNrobot}. The pooling layers after the first dilated convolution layer are removed (or stride set to 1) in these models. The proposed eConv and ePool layers utilize all pixels in the receptive field. A recent work removed the last two max pooling layers and increased last few convolution kernel size to $5 \times 5$ and $25 \times 25$~\cite{lr:Liu2015_ICCV_DPN}. Large convolution kernels are, however, computationally expensive.

If one changes the last pooling layer's stride to 1 (or by shifting and merging several pooling results), both denser predictions and equivalent nonlinear mappings are achieved~\cite{vi:Sermanet2014_Overfeat,lr:Pinheiro2014_RCNN,lr:Wu2016_SemanticSeg}. However, this strategy maintains equivalence only if there is no convolution or pooling layer after the changed pooling layer. Hence, the increasing in the number of predictions is small (usually $2^2=4$ or $3^2=9$). The proposed eConv and ePool layers can replace all (or any subset) of classic convolution and pooling layers in a CNN model, so long as the equivalent stride is set accordingly. If proper padding is added, eConv and ePool can produce prediction at every pixel location (i.e., $500 \times 500$ predictions for a $500 \times 500$ input image). However, to reduce the memory and CPU/GPU usage, we use classic convolution and pooling in the first few layers, and change later ones to eConv and ePool.

An early work in medical image segmentation implemented dense equivalent mappings~\cite{vi:Giusti2013}. The fast dense scan in~\cite{vi:Giusti2013} was implemented by generating multiple \emph{fragments} after every pooling layer, in which $s^2$ fragments are generated for a pooling layer with stride $s$. The network in~\cite{vi:Giusti2013} had 4 pooling layers and $s=2$; hence, 256 fragments were generated in the end. The strategy in~\cite{vi:Giusti2013} supports forward computation, but not error back propagation. The proposed eConv and ePool layers not only generate dense predictions,  but also support back propagation to learn better parameters.

\section{Experimental results} \label{sec:exp}

We use the proposed eConv and ePool layers to replace classic convolution and pooling layers in various CNN models. Because of the equivalence, the new model is initialized with the parameters of the original model. This initialization is also the final model in unsupervised tasks. Fine-tuning is used in supervised learning tasks to learn better parameters for the CNN. We call the new network the \texttt{dense} net. When we assign the dense net's parameters to the original network, we call it the \texttt{dpoa} (\underline{d}ense net's \underline{p}aramters, \underline{o}riginal net's \underline{a}rchitecture) net.

We show that dense nets are effective in various tasks, including at least semantic segmentation, object localization, recognition, and age estimation. We mainly use the FCN or VGG-16 model as the baseline CNN network, and present the empirical validation results in this section.

\subsection{Semantic segmentation}

Semantic segmentation is a natural application of the proposed eConv and ePool layers. A semantic segmentation system needs to predict the semantic category (e.g., horse, person, bicycle etc.) of every pixel in the input image. Hence, dense prediction is a desired property for such systems.

The baseline CNN we adopt is the fully convolutional network (FCN)~\cite{vi:Long2015_FCN}. The FCN-32s model outputs $16 \times 16$ predictions for a $500 \times 500$ input image, which is upsampled to predict for every pixel location. The number 32 means the upsampling ratio. By a skip connection between the fourth pooling layer (\texttt{pool4}) and the output, the prediction resolution becomes $34 \times 34$, which is called the FCN-16s model. Similarly, a skip connection from \texttt{pool3} leads to the FCN-8s model, whose output is the densest ($70 \times 70$) and the most accurate within the FCN family.

The proposed \texttt{dense} net changes the \texttt{pool3} layer's stride to 1 and converts subsequent convolution and pooling layers to eConv and ePool, respectively. Hence, the number of predictions in the \texttt{dense} net is 64($=8 \times 8$) times as many as that in FCN. As in FCN, we use \texttt{dense-$x$s} to denote the net learned with different skip connections, $x \in \{8,16,32\}$. The parameters in a \texttt{dense-32s} net are initialized from FCN-32s, and fine-tuned using the training set (e.g., Pascal VOC in this section.) The \texttt{dense-16s} net is fine-tuned from \texttt{dense-32s}, and \texttt{dense-8s} is fine-tuned from \texttt{dense-16s}.

\paragraph{VOC 2011.} We first tested on the Pascal VOC 2011 dataset, using the \emph{same} training and validation images as the original FCN~\cite{vi:Long2015_FCN}. The training set include 8825 images (1112 from VOC2011 and the rest are extended ones from~\cite{lr:Hariharan2011}); the validation set has 736 images. The mean class-wise intersection over union (IoU) score is the main metric to evaluate semantic segmentation results, and a higher IoU means better segmentation. The background is considered as a class in computing the mean IoU. We tested FCN, \texttt{dense} and \texttt{dpoa} nets on both the validation set and the VOC 2011 test server.

\paragraph{VOC 2012.} We also tested on the Pascal VOC 2012 dataset using the same training and validation images as those in~\cite{vi:Zheng2015_RNN_seg}. The training set includes 11685 images (1464 from VOC 2012 and the rest are extended ones from~\cite{lr:Hariharan2011}); the validation set has 346 images. We tested both FCN and \texttt{dense} nets on both the validation set and the VOC 2012 test server. Results on both VOC 2011 and VOC 2012 are shown in Table~\ref{tbl:VOC}.

\begin{table}
 \caption{IoU comparison on Pascal VOC 2011 and VOC 2012~\cite{vi:Everingham2015_VOC}.}
 \centering
 \small
 \label{tbl:VOC}
 \subfloat[2011 validation] { 
   \label{tbl:VOC_2011_val}
   \begin{tabular}{@{\,}l@{\,}|@{\,}r@{\,}|@{\,}l@{\,}|@{\,}r@{\,}}
    \hline
    & IoU & & IoU \\ \hline
    FCN-32s & 59.4\% & \texttt{dense-32s} & 62.4\% \\ \hline
    FCN-16s & 62.4\% & \texttt{dense-16s} & 64.8\% \\ \hline
    FCN-8s & 62.7\% & \texttt{dense-8s} & 65.3\% \\ \hline
   \end{tabular}
 }
 \hspace{12pt}
 \subfloat[2011 test] {
  \label{tbl:VOC_2011_test}
  \begin{tabular}{@{\,}l@{\,}|@{\,}r@{\,}}
   \hline
   & IoU \\ \hline
   FCN-8s & 62.7\%  \\ \hline
   \texttt{dense-8s} & 66.7\% \\ \hline
   \texttt{dpoa-8s} & 65.6\%  \\ \hline
  \end{tabular}
 }
 \hspace{12pt}
 \subfloat[2012 validation] {
  \label{tbl:VOC_2012_val}
  \begin{tabular}{@{\,}l@{\,}|@{\,}r@{\,}}
   \hline
   & IoU \\ \hline
   FCN-8s & 61.3\%  \\ \hline
   \texttt{dense-8s} & 66.2\%  \\ \hline
   \texttt{dpoa-8s} & 63.9\% \\ \hline
  \end{tabular}
 }
 \hspace{12pt}
 \subfloat[2012 test] {
  \label{tbl:VOC_2012_test}
  \begin{tabular}{@{\,}l@{\,}|@{\,}r@{\,}}
   \hline
   & IoU \\ \hline
   FCN-8s & 62.2\%  \\ \hline
   \texttt{dense-8s} & 68.4\% \\ \hline
   \texttt{dpoa-8s} & 66.5\%  \\ \hline
  \end{tabular}
 }
\end{table}

\paragraph{The \texttt{dense} nets are more accurate than baseline FCNs.} As shown in Tables~\ref{tbl:VOC_2011_val} to~\ref{tbl:VOC_2012_test}, the proposed \texttt{dense} nets are more accurate than their corresponding FCN nets. The smallest improvement is 2.4\%, between FCN-16s and \texttt{dense-16s} on the VOC 2011 validation set. On the test set and VOC 2012, the improvement is much larger. For example, the mean IoU of \texttt{dense-8s} is 6.2\% higher than that of FCN-8s on the VOC 2012 test set!

\paragraph{The \texttt{dpoa} net is a good compromise between speed and accuracy.} The \texttt{dpoa} net has exactly the same architecture and speed as the baseline FCN net. As shown in Tables~\ref{tbl:VOC_2011_test} and~\ref{tbl:VOC_2012_test}, the \texttt{dpoa} net has 2.8\% and 4.3\% higher mean IoU than baseline FCN-8s on the VOC 2011 and VOC 2012 test sets, respectively. Its accuracy is much closer to the \texttt{dense} net than FCN. Hence, when fast testing speed is required, the proposed \texttt{dpoa} net is very useful.

\paragraph{The increase of training examples helps \texttt{dense} and \texttt{dpoa}.} We show the per-category IoU on the VOC 2012 test set in Table~\ref{tbl:VOC12test_perclass}. In \emph{every} category, we always observe the same ranking in terms of IoU: \texttt{dense}>\texttt{dpoa}>FCN, which further supports the effectiveness of both proposed networks. However, we observe that the improvement is small in categories with thin structures (e.g., bicycle and chair), but is large for objects with bulk structures (e.g., cow and potted-plant). We conjecture that the higher IoU of \texttt{dense} is because more training examples and labels are used due to the increased number of predictions. The performance of \texttt{dpoa} also supports this conjecture. It has the same architecture as FCN, hence its higher IoU means that its parameters are better than that in FCN, which is the consequence of training with more pixels (denser predictions) and their labels.

\begin{table}
 \caption{Pascal VOC 2012 test set IoU comparison. The baseline CNN is FCN-8s.} \label{tbl:VOC12test_perclass}
 \centering
 \scriptsize
 \begin{tabular}{@{\,\,}c@{\,\,}|*{20}{|@{\,}r@{\,}}||@{\,\,}c@{\,\,}}
  \hline
    &  plane &  bike &  bird &  boat &  bott &  bus &  car &  cat &  chair &  cow &  tbl &  dog &  horse &  motor &  person &  plant &  sheep &  sofa &  train &  tv &  mean \\ \hline
  FCN-8s & 76.8 & 34.2 & 68.9 & 49.4 & 60.3 & 75.3 & 74.7 & 77.6 & 21.4 & 62.5 & 46.8 & 71.8 & 63.9 & 76.5 & 73.9 & 45.2 & 72.4 & 37.4 & 70.9 & 55.1 & 62.2 \\ \hline
  \texttt{dense} & \textbf{83.0} & \textbf{38.6} & \textbf{74.1} & \textbf{58.6} & \textbf{65.7} & \textbf{83.2} & \textbf{81.5} & \textbf{80.7} & \textbf{29.2} & \textbf{73.9} & \textbf{53.8} & \textbf{74.8} & \textbf{73.0} & \textbf{79.3} & \textbf{78.3} & \textbf{56.3} & \textbf{80.2} & \textbf{46.2} & \textbf{72.7} & \textbf{61.0} & \textbf{68.4} \\ \hline
  \texttt{dpoa} & 79.4 & 35.0 & 72.4 & 55.0 & 65.5 & 82.3 & 78.8 & 79.4 & 27.1 & 71.3 & 53.1 & 74.3 & 70.8 & 77.9 & 75.9 & 53.5 & 79.0 & 42.9 & 71.3 & 59.3 & 66.5 
\\ \hline
 \end{tabular}
\end{table}

\begin{figure}
 \centering
 \subfloat[Input] { \includegraphics[width=0.18\textwidth]{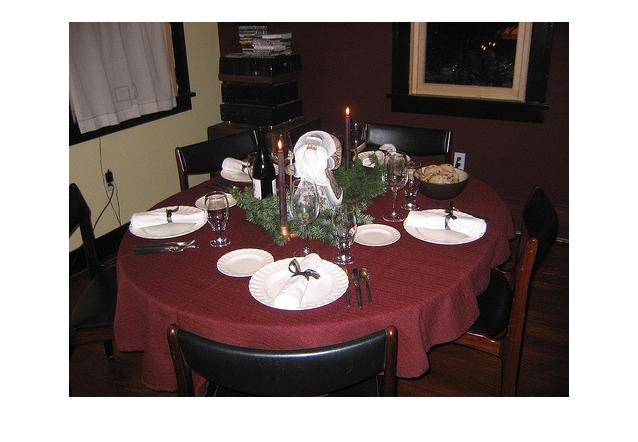} \label{fig:SSResult_INPUT} }
 \subfloat[Groundtruth] { \includegraphics[width=0.18\textwidth]{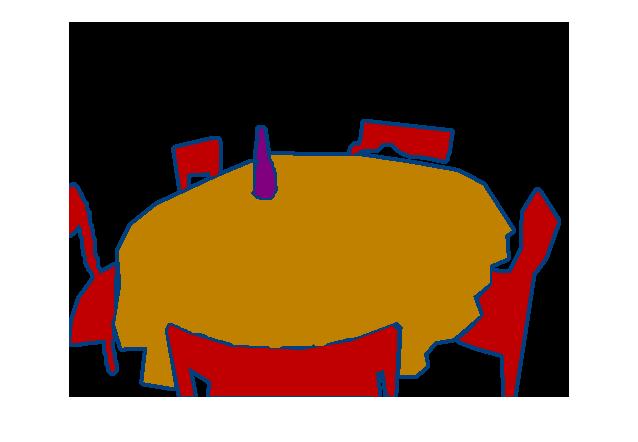} \label{fig:SSResult_GT} }
 \subfloat[FCN-8s] { \includegraphics[width=0.18\textwidth]{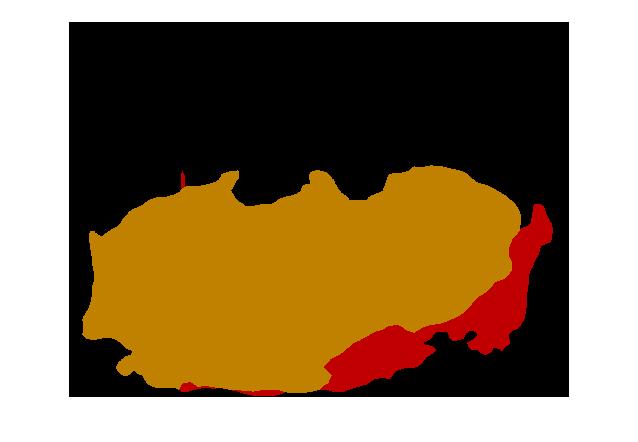} \label{fig:SSResult_FCN} }  \\
 \subfloat[\texttt{dense-8s}] { \includegraphics[width=0.18\textwidth]{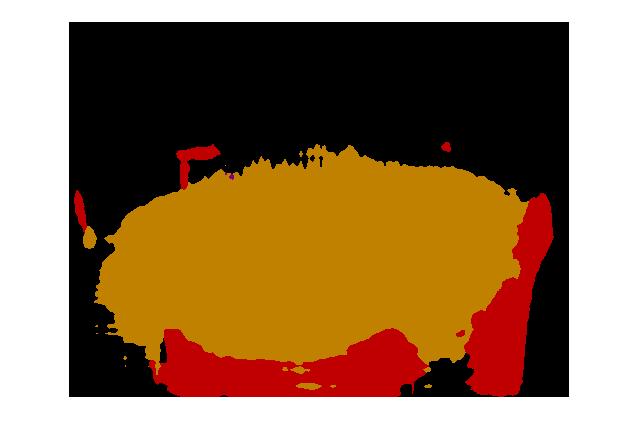} \label{fig:SSResult_DENSE} }
 \subfloat[\texttt{dpoa-8s}] { \includegraphics[width=0.18\textwidth]{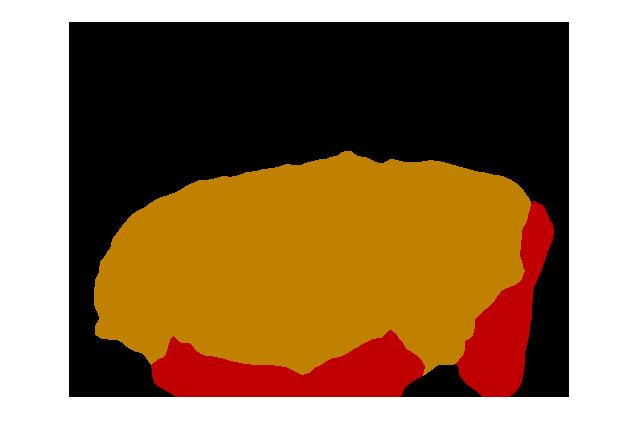} \label{fig:SSResult_DPOA} }
 \caption{Example semantic segmentation results of different models.} \label{fig:SSResult}
\end{figure}

\paragraph{The \texttt{dense} and \texttt{dpoa} nets may be further improved by adding CRF and more training images.} The conditional random field (CRF) has been shown to boost the accuracy of semantic segmentation methods by a large margin~\cite{vi:Chen2015_DeepLab,lr:Wu2016_SemanticSeg,lr:Yu2016_DilationConv,vi:Zheng2015_RNN_seg}. Although we did not experiment with adding CRF on top of the proposed nets, we expect this combination will lead to higher segmentation accuracy. As shown in Figure~\ref{fig:SSResult}, \texttt{dense-8s} detects more details of the chairs than FCN-8s or \texttt{dpoa-8s}. However, it also contains more non-smooth predictions, such as the yellow pixels within the group of red pixels (i.e., predicting a table inside a chair). CRF is an expert in correcting this kind of errors. The usage of additional training data, e.g., MS COCO (\url{http://mscoco.org/}) should be useful, too.

\begin{figure}
 \centering
 \subfloat[VGG-16 net] { \includegraphics[scale=0.25]{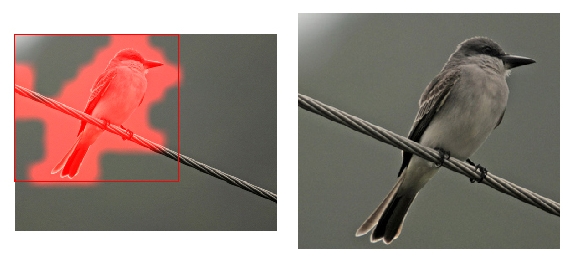} \label{fig:cub_vgg} }
 \hspace{3pt} \vrule \hspace{3pt}
 \subfloat[\texttt{dense} net] { \includegraphics[scale=0.25]{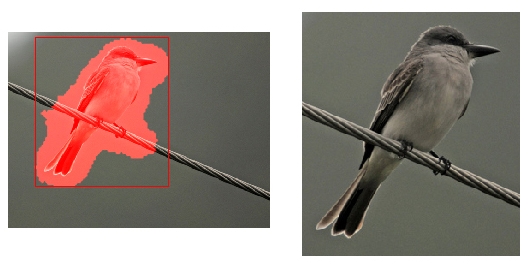} \label{fig:cub_dense} }
 \caption{Unsupervised object localization using the VGG-16 net (\ref{fig:cub_vgg}) and \texttt{dense} net (\ref{fig:cub_dense}).} \label{fig:cub}
\end{figure}

\subsection{Unsupervised object localization}

eConv and ePool layers are effective in unsupervised object localization, too. Wei et al.~\cite{me:Wu2016_SCDA} proposed an unsupervised method to localize the main object in fine-grained datasets, e.g., the CUB200-2011 dataset~\cite{vi:Wah2011_CUB}. They showed that the sum of \texttt{pool5} output ($7 \times 7 \times 512$, across the channel axis) of VGG-16 is a good indicator to localize the main object. The sum ($7 \times 7$ in size) is truncated and converted to a binary mask using their average value as the threshold, and resized to the input image size as a mask to localize the object. Figure~\ref{fig:cub_vgg} shows one example of the localization, including the input image and the mask (shown in red), and the localized object in the bounding box computed from the mask (enlarged in the figure). Using the 50\% IOU criterion, the localization accuracy is 76.09\% in our implementation (76.79\% in~\cite{me:Wu2016_SCDA}) on the CUB200-2011 dataset.

We change the stride of \texttt{pool4} to 1 and change subsequent layers to eConv and ePool in the VGG-16 net. Since this is an unsupervised object localization method, no fine-tuning is performed. As shown in Figure~\ref{fig:cub_dense}, this \texttt{dense} net can localize the bird with tighter boundaries. The localization accuracy is 79.06\% using the 50\% IOU criterion. The dense net has achieved a 3\% improvement over the VGG-16 net. Because no new parameters are obtained, a \texttt{dpoa} net is not applicable for this task. As in the semantic segmentation, eConv and ePool layers provide denser predictions than the original network, hence more accurate localization is obtained, although the same parameters are used in the \texttt{dense} net and the VGG-net.

\subsection{Image categorization}

Image categorization is a classic application of CNN~\cite{lr:LeCun1998,vi:Krizhevsky2012}. We tested the \texttt{dense} net on the scene categorization dataset \texttt{indoor}~\cite{pl:Quattoni2009}, which has 67 scene classes (roughly 80 training and 20 testing images per category). The baseline net is VGG-16 fine-tuned on the training images. The input image is resized to $256 \times 256$ and random crops of size $224 \times 224$ are used for training. $256 \times 256$ images are used directly for testing, yielding $2 \times 2$ probability estimates. The final prediction is based on the average of them. In the \texttt{dense} net, we change the \texttt{pool4} stride to 1 and subsequent layers to eConv and ePool, which generates $31 \times 31$ probability estimates. We add a global average pooling layer to turn them into one estimate. The \texttt{dense} net is initialized with the baseline model's parameters and fine-tuned using the training images.

On the \texttt{indoor} dataset, the fine-tuned VGG-16 net's accuracy is 73.06\%, \texttt{dense} net has 73.51\% accuracy, and the \texttt{dpoa} net achieves 74.02\% accuracy. The \texttt{dense} net has higher accuracy than VGG-16, but the \texttt{dpoa} net has the highest among them. Because image categorization is not a pixel labeling task, we conjecture that averaging a large number of dense predictions is not as effective as directly applying the better parameters to the original CNN architecture.

We also evaluated the testing speed on the \texttt{indoor} dataset. The running time of VGG-16, \texttt{dpoa} and \texttt{dense} are 44ms, 44ms and 132ms per image, respectively. As expected, the testing speed of \texttt{dpoa} is as fast as that of VGG-16. As for \texttt{dense}, although it produces 240 times more predictions, its speed is only 2 times slower than that of the baseline.

\subsection{Apparent age estimation}

We also tested the \texttt{dense} net on the apparent age estimation problem, which tries to estimate the age of a person based on a single picture. The dataset is the ChaLearn 2015 competition data, which has 2476 training and 1136 validation images.\footnote{\url{http://gesture.chalearn.org/2015-looking-at-people-iccv-challenge}} We start from the VGG-Face model~\cite{vi:Parkhi2015_VGGFace}. Given an input image, the face within it is detected, cropped and aligned to the frontal pose. It is then resized to $224 \times 224$. The groundtruth age $a$ is converted into the range $[-1 \; +1]$ by $\frac{a-45}{45}$. We add a $\tanh$ and an $\ell_1$ loss layer to VGG-Face, and use the ChaLearn training images to fine-tune and get the baseline CNN model (which we still call VGG-Face). To train the \texttt{dense} net, we change the stride of VGG-Face's \texttt{pool4} layer to 1 and change subsequent layers to eConv and ePool. The last fully connected layer will output $27 \times 27$ predictions. We add a global average pooling layer to summarize them, and also add a $\tanh$ and an $\ell_1$ loss layer. The \texttt{dense} net is initialized with the baseline model's parameters and fine-tuned using the training images.

The evaluation uses the validation set, and the metric is mean absolute error (MAE) (i.e., number of years difference between the groundtruth and the prediction). The MAE of VGG-Face, \texttt{dense} and \texttt{dpoa} are 6.82, 5.28 and 6.19 years, respectively. Both \texttt{dense} and \texttt{dpoa} have smaller errors than the baseline net. This time \texttt{dense} has a large winning margin over \texttt{dpoa} (0.91 year, or 15\% relative improvement). Our conjecture is: because the face alignment cannot be perfect, the ensemble of a lot of predictions has an advantage in compensating for the inaccurate alignments.

\section{Conclusions and discussions} \label{sec:conclusion}

In this paper, we introduced two layers for CNN: equivalent convolution (eConv) and equivalent pooling (ePool). These layers can turn many existing CNN models into new networks that output dense predictions. The same (equivalent) nonlinear mapping is applied to different receptive fields in the input image, including those already computed by the original model and many more.

The equivalence allows learning the dense model starting from existing CNN models, and more importantly, allows transferring better parameters in the dense model back to the baseline CNN model and improving its accuracy. This inverse transfer allows both fast speed of the original CNN net and higher accuracy. The proposed layers are not only useful for applications requiring dense pixel labeling, but many others. Experiments on semantic segmentation, unsupervised object localization, image categorization and age estimation verified the effectiveness of the proposed layers.

An obvious future research direction is to test the proposed eConv and ePool layers in more CNN architectures, e.g., CNN+CRF in semantic segmentation (e.g.,~\cite{lr:Lin2015_NIPS_CRF}) and deeper CNN architectures such as the ResNet~\cite{vi:He2016_ResNet} and GoogLeNet~\cite{vi:Szcgcdy2015_GoogLeNet}. The proposed dense prediction layers are also expected to improve other tasks requiring dense pixel labeling, such as object detection without resorting to region proposals, depth estimation from single color image, saliency detection, scene understanding for automatic (autonomous) driving and intrinsic image decomposition. It is also potentially useful for learning recurrent CNN (e.g.,~\cite{lr:Liang2015_NIPS}).

\small
\bibliographystyle{abbrv}
\bibliography{abbr_s,BibAll}

\end{document}